\documentclass[conference]{IEEEtran}
\ifCLASSINFOpdf
\else
\fi

\usepackage{times}
\usepackage{epsfig}
\usepackage{graphicx}
\usepackage{amsmath}
\usepackage{amssymb}

\usepackage{amsfonts}%
\usepackage{stackrel}
\usepackage{amsthm}
\usepackage{algorithmic}
\usepackage{epstopdf} % to include .eps graphics files with pdfLaTeX
\usepackage{float}
\usepackage{booktabs}
\usepackage{subfigure}

\newtheorem{thm2}{Theorem}
\newtheorem{mydef}{Definition}
\newtheorem{myprob}{Problem}

\begin{document}
\title{Submodular Optimization for Efficient Semi-supervised Support Vector Machines}

\author{\IEEEauthorblockN{Wael Emara and Mehmed Kantardzic}
\IEEEauthorblockA{Computer Engineering and Computer Science Department\\
University of Louisville, 
Louisville, Kentucky 40292\\
Email: waemar01@cardmail.louisville.edu, mmkant01@louisville.edu}}


%Telephone: (800) 555--1212, Fax: (888) 555--1212}


\maketitle

\begin{abstract}
%\boldmath
In this work we present a quadratic programming approximation of the Semi-Supervised Support Vector Machine (S3VM) problem, namely approximate QP-S3VM, that can be efficiently solved using off the shelf optimization packages. We prove that this approximate formulation establishes a relation between the low density separation and the graph-based models of semi-supervised learning (SSL) which is important to develop a unifying framework for semi-supervised learning methods. Furthermore, we propose the novel idea of representing SSL problems as submodular set functions and use efficient submodular optimization algorithms to solve them. Using this new idea we develop a representation of the approximate QP-S3VM as a maximization of a submodular set function which makes it possible to optimize using efficient greedy algorithms. We demonstrate that the proposed methods are accurate and provide significant improvement in time complexity over the state of the art in the literature. 
\end{abstract}


% \fi
\IEEEpeerreviewmaketitle

\section{Introduction}
The recent advances in information technology imposes serious challenges on traditional machine learning algorithms where classification models are trained using labeled samples. Data collection and storage nowadays has never been easier and therefore using such enormous volumes of data to infer reliable classification models is of utmost importance. Meanwhile, labeling entire data sets to train classification models is no longer a valid option due to the high cost of experienced human annotators. Despite the recent efforts to make annotation of large data sets cheap and reliable by using online workforce, the collected labeled data can never keep up with the cheap collection of unlabeled data.

Semi-supervised learning (SSL) handles this issue by utilizing large amount of unlabeled samples, along with labeled samples to build better performing classifiers. Two assumptions form the basis for the usefulness of unlabeled samples in discriminative SSL methods: the cluster assumptions and the smoothness assumption \cite{zhu05survey}. Although both assumptions use the idea that samples that are close under some distance metric should assume the same label, they inspire different categories of SSL algorithms, namely low density separation methods (for 
the cluster assumption) and graph-based methods (for the smoothness 
assumption). In the low density separation methods the unlabeled samples are used to better estimate the boundaries or each class. The graph-based methods use labeled and unlabeled samples to construct a graph representation of the data set where information is then propagated from the labeled samples to the unlabeled samples through the dense regions of the graph, a process known as {\em{label propagation}} \cite{SSLHarmonic}.

The practical success and the theoretical robustness of large margin methods in general and specially Support Vector Machines (SVM) has drawn a lot of attention to Semi-Supervised Support Vector Machines (S$^3$VM) \cite{bib_5369}. However the problem is challenging due to the non-convexity of the objective function. In this paper we propose an approximate-S$^3$VM formulation that will result in a standard quadratic programming problem, namely approximate QP-S$^3$VM, that can be solved directly using off the shelf optimization packages. One important aspect of the proposed formulation is that it uncovers a connection between the S$^3$VM, as a low density separation method, and the graph based algorithms which is a helpful step towards a unifying framework for SSL \cite{LDS_NIPS_06}. Furthermore, we present a new formulation of loss based SSL problems. The new formulation represents SSL problems as set functions and use the theory of {\em{submodular set functions}} optimization to solve them efficiently. Specifically, we present a submodular set function that is equivalent to the proposed approximate QP-S$^3$VM and solve it efficiently using a greedy approach that is well established in optimizing submodular functions \cite{Nemhauser1}.

Section \ref{Sec:Prelim} provides preliminaries of S$^3$VM and the notations used throughout the paper. The proposed approximate QP-S$^3$VM is detailed in Section \ref{Sec:QP-S$^3$VM}. In Section \ref{Sec:SubmodOpt} we present the submodular formulation of the approximate QP-S$^3$VM. Experimental results are provided in Section \ref{Sec:Exp}, followed by the conclusion in Section \ref{Sec:Conclusion}.

\subsection{Preliminaries}
\label{Sec:Prelim}
Semi-supervised learning uses partially labeled data sets $\mathcal{L}\cup\mathcal{U}$ where $\mathcal{L}=\{({\bf{x}}_{i},y_{i})\}$ and $\mathcal{U}=\{{\bf{x}}_{j}\}$, ${\bf{x}}\in \mathbb{R}^n$, and $y_i\in \{ +1,-1\}$. Throughout this paper we use $i$ and $j$ as indices for labeled and unlabeled samples, respectively.

The major body of work on  S$^3$VM is based on the idea of solving a standard SVM while treating unknown labels as additional variables \cite{bib_5369}. The semi-supervised learning problem is to find the solution of
\begin{equation}
\label{S3VMEq}
\begin{aligned}
\stackrel[\hspace{1 mm}{\bf{w}},y_j]{}{min}\mathcal{J}({\bf{w}},{{y}}_j)= \hspace{1 mm} \frac{1}{2}\| \textbf{w} \|^2+&C\sum_{i\in\mathcal{L}} \ell_{l} ({\bf{w}}, ({\bf{x}}_{i},y_{i}))\\&+ C^\ast\sum_{j\in\mathcal{U}}\ell_{u}({\bf{w}},{\bf{x}}_{j})
\end{aligned}
\end{equation}
where the loss functions for unlabeled samples $\ell_u$ and labeled samples $\ell_l$ are defined as follows:
\begin{equation}
\label{S3VMLossEq}
\ell_{u}({\bf{w}},({\bf{x}}_j,y_j))=\stackrel[y_{j}\in\{-1,+1\}]{}{max}\{0,1-y_{j}(\langle{\bf{w}},{\bf{x}}_j\rangle+b)\}
\end{equation}
\begin{equation}
\label{S3VMLossEq2}
\ell_{l}({\bf{w}},({\bf{x}}_i,y_i))=\stackrel[]{}{max}\{0,1-y_{i}(\langle{\bf{w}},{\bf{x}}_i\rangle+b)\}
\end{equation}
The solution of Eqn.\eqref{S3VMEq} will result in finding the optimal separating hyperplane ${\bf{w}}$ and the labels assigned to the unlabeled samples $y_j$. The loss over labeled and unlabeled samples is controlled by two parameters $C$ and $C^*$, which reflect the confidence in the labels $y_i$ and the cluster assumption, respectively.

Algorithms that solve Eqn.\eqref{S3VMEq} can broadly be divided into combinatorial and continuous optimization algorithms. In continuous optimization algorithms, for a given fixed ${\bf{w}}$, the optimal $y_j$ are simply obtained by $sgn(\langle{\bf{w}},{\bf{x}}_j\rangle+b)$. The problem then comes down to a continuous optimization problem in ${\bf{w}}$. On the other hand, in combinatorial optimization algorithms, for given $y_j$, the optimization for ${\bf{w}}$ is a standard SVM problem. Therefore, if we define a function $\mathcal{I}$($y_j$) such that
\begin{equation}
\label{CombinatorialS3VM}
\mathcal{I}(y_j)=\stackrel[{\bf{w}}]{}{min}\mathcal{J}({\bf{w}},{{y}}_j)
\end{equation}
the problem will be transformed to minimizing $\mathcal{I}(y_j)$ over a set of binary variables where each evaluation of $\mathcal{I}(y_j)$ is a standard SVM optimization problem \cite{DBLP:conf/nips/ChapelleSK06,citeulike:384511,DBLP:conf/icml/SindhwaniKC06},
\begin{equation}
\label{CombinatorialS3VM}
\stackrel[y_j]{}{min}\mathcal{I}(y_j).\end{equation}   
Solving Eqn.\eqref{S3VMEq} may lead to degenerate solutions where all the unlabeled samples are assigned to one class. This is usually handled in the literature by enforcing a {\em{balancing constraint}} which makes sure that a certain ratio $r$ of the unlabeled samples are assigned to class +1 \cite{bib_5369}. 
\section{Quadratic Programming Approximation of S$^3$VM (QP-S$^3$VM)}
\label{Sec:QP-S$^3$VM}
In Eqn.\eqref{CombinatorialS3VM} the combinatorial formulation of S$^3$VM optimizes for the labels $y_j$ that minimize the loss associated with each unlabeled sample. To overcome the hard combinatorial problem, the loss of setting $y_j=1$, denoted by $\ell{_j^+}$, is assigned a new variable $p_j$, where $0\leq p_j\leq 1$. This variable indicates the probability that the $y_j=1$ is correct. Similarly, the loss of setting $y_j=-1$, denoted by $\ell{_j^-}$, is given by the probability $1-p_j$. The balancing constraint will have the form $\sum_{j\in\mathcal{U}}p_j=r|\mathcal{U}|$.
{This modified formulation has the following form \cite{DBLP:conf/icml/SindhwaniKC06,Wang:2009:ELM:1577069.1577094}:

\begin{myprob}
\label{ContinFormOfCombS3VM}
Continuous optimization formulation of the combinatorial S$^3$VM problem. 
\begin{equation}
\label{S3VMwithPVariables}
\begin{aligned}
&\stackrel[{\bf{P}}]{}{argmin}~\stackrel[{\bf{w}}]{}{min} 
\mathcal{J}({\bf{w}},{\bf{P}})=\frac{1}{2}\|{\bf{w}}\|^2+C\sum_{i\in\mathcal{L}}\zeta_i+C^*\sum_{j\in\mathcal{U}}p_j\ell^+_j\\&~~~~~~~~~~~~~~~~~~~~~+C^*\sum_{j\in\mathcal{U}}(1-p_j)\ell^-_j\\
&\begin{aligned}
 \text{subject to } ~~~~~~~y_i[\langle{\bf{w}},{\bf{x}}_i\rangle+b]\geq 1-\zeta_i\\
 \langle{\bf{w}},{\bf{x}}_j\rangle+b\geq 1-\ell^+_j\\ 
 -\langle{\bf{w}},{\bf{x}}_j\rangle-b\geq 1-\ell^-_j\\
 \zeta_i\geq 0,\ell^+_j\geq 0,\ell^-_j\geq 0\\ 0\leq p_j\leq 1, \stackrel[j\in\mathcal{U}]{}{\sum}p_j=r|\mathcal{U}|
\end{aligned}
\end{aligned}
\end{equation}
\end{myprob}

Now that the problem has been simplified from being combinatorial in $y_i$, $y_j\in\{+1,-1\}$, to being continuous in $p_j$, $p_j\in[0,1]$, we proceed to find the dual form. Deriving the {\em{Lagrangian}} of the continuous formulation in Problem \ref{ContinFormOfCombS3VM} and applying the {\em{Karush-Kuhn-Tucker}} conditions to it, the obtained dual form is presented in Problem \ref{DualProb}.
\begin{myprob}
\label{DualProb}
Dual form of $\stackrel[{\bf{w}}]{}{min} 
\mathcal{J}({\bf{w}},{\bf{P}})$ in Problem \ref{ContinFormOfCombS3VM}.
\begin{eqnarray}
\stackrel[{\bf{A}},{\bf{B}},{\bf{\Gamma}}]{}{max}{\bf{\mathcal{I}}_{Dual}}
\end{eqnarray}
where
\begin{equation}
\begin{aligned}
{\bf{\mathcal{I}}_{Dual}}={\bf{A'1}}_{|\mathcal{L}|}+({\bf{\Gamma}}+{\bf{B}})'{\bf{1}}_{|\mathcal{U}|}-\frac{1}{2}({\bf{A}}\circ {\bf{Y}})'{\bf{K}_{ll}}({\bf{A}}\circ {\bf{Y}})\\
-\frac{1}{2}({\bf{\Gamma}}-{\bf{B}})'{\bf{K}_{uu}}({\bf{\Gamma}}-{\bf{B}})
-({\bf{A}}\circ{\bf{Y}})'{\bf{K}_{lu}}({\bf{\Gamma}}-{\bf{B}})
\end{aligned}
\end{equation}
%$\text{subject to}$
\begin{eqnarray*}
\begin{aligned}
\text{subject to} &~~~{\bf{0}}\leq{\bf{A}}\leq C{{\bf{1}}_{|\mathcal{L}|}}\\
&~~~{\bf{0}}\leq{\bf{\Gamma}}\leq C^*\bf{P}\\
&~~~{\bf{0}}\leq{\bf{B}}\leq C^*({\bf{1}}_{|\mathcal{U}|}-{\bf{P}})
\end{aligned}
\end{eqnarray*}
$
where\\
\begin{aligned}
&{\bf{1}}_{|\mathcal{L}|}\text{: A ones vector of length } |\mathcal{L}|.\text{ Similarly is }{\bf{1}}_{|\mathcal{U}|}.\\
&\alpha_i \text{:Lagrangian Multiplier of labeled loss constraint }\zeta_i.\\
&\gamma_j \text{:Lagrangian Multiplier of unlabeled loss constraint }\ell^+_j.\\
&\beta_j \text{:Lagrangian Multiplier of unlabeled loss constraint }\ell^-_j.\\
&{\bf{A}}'=[\alpha_1,\dots,\alpha_{|\mathcal{L}|}], {\bf{B}}'=[\beta_1,\dots,\beta_{|\mathcal{U}|}], {\bf{\Gamma}}'=[\gamma_1,\dots,\gamma_{|\mathcal{U}|}]\\
&{\bf{P}}'=[p_1,\dots,p_{|\mathcal{U}|}],~~{\bf{Y}}'=[y_1,\dots,y_{|\mathcal{L}|}],\\
&{\bf{K_{ll}}}={\bf{K}}_{i,i'}~\forall i,i'\in\mathcal{L},~~~{\bf{K_{uu}}}={\bf{K}}_{j,j'}~\forall j,j'\in\mathcal{U},\\&{\bf{K_{lu}}}={\bf{K}}_{i,j}~\forall i\in\mathcal{L},j\in\mathcal{U}.
\end{aligned}
$

\end{myprob} 
    
Using the derived dual form in Problem \ref{DualProb}, we propose an approximate optimization based on minimizing an upper bound of $max_{{\bf{A}},{\bf{B}},{\bf{\Gamma}}}\bf{\mathcal{I}}_{Dual}$. The proposed upper bound is specified in the following theorem.
\begin{thm2}
Proposed upper bound for $max_{{\bf{A}},{\bf{B}},{\bf{\Gamma}}}\bf{\mathcal{I}}_{Dual}$:
\label{UpperBoundThm}
\begin{equation}
\begin{aligned}
\stackrel[{\bf{A}},{\bf{B}},{\bf{\Gamma}}]{}{max}{\bf{\mathcal{I}}_{Dual}}
\leq {\bf{\mathcal{I}}}({\bf{w^*}})+C^*|\mathcal{U}|+{\bf{\mathcal{M}}_1}+{\bf{\mathcal{M}}_2}
\end{aligned}
\end{equation}
%$where$
\begin{equation}
\begin{aligned}
\text{where}~~~{\bf{\mathcal{I}}}({\bf{w^*}}) &= \stackrel[{\bf{w}}]{}{min}\frac{1}{2}\|{\bf{w}}\|^2+C\sum_{i\in\mathcal{L}}\zeta_i\\
{\bf{\mathcal{M}}_1}&=\frac{1}{2}{C^*}^2({\bf{1}_{|\mathcal{U}|}}-{\bf{P}})'{\bf{K}_{uu}}{\bf{P}}\\
{\bf{\mathcal{M}}_2}&=CC^*{\bf{Y}}'{\bf{K}_{lu}}({\bf{1}_{|\mathcal{U}|}}-{\bf{P}})\\
\end{aligned}
\end{equation}
\end{thm2}
\begin{proof} See the appendix.
\end{proof}
   
Examining the upper bound in Theorem \ref{UpperBoundThm}, ${\bf{\mathcal{I}}}({\bf{w^*}})$ is the objective function value of optimizing a standard supervised SVM on the labeled samples $\mathcal{L}$. Therefore, it is constant as well as the term $C^*|\mathcal{U}|$. The rest of the upper bound, ${\bf{\mathcal{M}}_1}+{\bf{\mathcal{M}}_2}$, is a function of ${\bf{P}}$. The optimal values of ${\bf{P}}$ are now obtainable through the following optimization problem.

\begin{myprob}
\label{QPS3VMProb}
Quadratic programming approximation of Semi-supervised Support Vector Machines (QP-S$^3$VM):
\begin{equation}
\label{QP-S$^3$VMObjective}
\begin{aligned}
\stackrel[{\bf{P}}]{}{min }\frac{1}{2}{C^*}^2{\bf{(1_{|\mathcal{U}|}-P)'{K}_{uu}P}}
+CC^*{{\bf{Y}}}'{{\bf{K}_{lu}}{\bf{(1_{|\mathcal{U}|}-P)}}}
\end{aligned}
\end{equation}
$\text{subject to}$
\begin{equation}
\begin{aligned}
{\bf{P}}'{\bf{1}_{|\mathcal{U}|}}=r|\mathcal{U}|,~~~~~ {\bf{0\leq P\leq 1}_{|\mathcal{U}|}}.
\end{aligned}
\end{equation}
Note: Equation \eqref{QP-S$^3$VMObjective} can be rewritten in the standard quadratic programming form as follows:
\begin{equation}
\label{QP-S$^3$VMObjectiveStandard}
\begin{aligned}
\stackrel[{\bf{P}}]{}{min }-\frac{1}{2}{C^*}^2{\bf{P'{K}_{uu}P}}
+(\frac{1}{2}{C^*}^2{\bf{1_{|\mathcal{U}|}}}{\bf{{K}_{uu}}}-CC^*{{\bf{Y}}}'{{\bf{K}_{lu}}}){\bf{P}}
%CC^*{{\bf{Y}}}'{{\bf{K}_{lu}}{\bf{(1_{|\mathcal{U}|}-P)}}}
\end{aligned}
\end{equation}
\end{myprob} 
The proposed approximate formulation is a quadratic programming problem in the variables $p_j$. In order to avoid trivial solutions to the problem where all the variables $p_j$ are $zero$. We add the constraint ${\bf{P}}'{\bf{1}}=r|\mathcal{U}|$ which makes sure that a certain ratio of the unlabeled samples, $r$, be assigned to class $+1$. %This constraint is known in the literature as the balancing constraint \cite{bib_5369}.

\subsection{QP-S$^3$VM Model Interpretation}

In this section we analyze the approximate model obtained in Problem \ref{QPS3VMProb}. This is necessary to ensure that the approximate model does not deviate from the original S$^3$VM problem. The first term in Eqn.\eqref{QP-S$^3$VMObjective} can be expanded as follows:
\begin{equation}
\begin{aligned}
\frac{1}{2}{C^*}^2&({\bf{1}_{|\mathcal{U}|}}-{\bf{P}})'{\bf{K}_{uu}}{\bf{P}}
\\=&\stackrel[\mathcal{Q}_1]{}{\underbrace{\frac{1}{2}{C^*}^2\!\!\!\!\!\!\stackrel[\substack{j,j'=\{1,\dots,|\mathcal{U}|\}\\j=j'}]{}{\sum}\!\!\!\!\!\![{\bf{K}_{uu}}]_{j,j'}p_{j'}(1-p_j)}}
\\&+\stackrel[\mathcal{Q}_2]{}{\underbrace{\frac{1}{2}{C^*}^2\!\!\!\!\!\!\stackrel[\substack{j=\{1,\dots,|\mathcal{U}|-1\}\\j'=\{j+1,\dots,|\mathcal{U}|\}}]{}{\sum}\!\!\!\!\!\![{\bf{K}_{uu}}]_{j,j'}(p_j+p_{j'}-2p_jp_{j'})}}
\end{aligned}
\end{equation}
As $\mathcal{Q}_1$ is negative quadratic in $p_j$, minimizing $\mathcal{Q}_1$ enforces the values of $p_j$ to be either $0$ or $1$. In other words, minimizing $\mathcal{Q}_1$ help making clear assignments of the labels to the unlabeled samples. To understand the implications of minimizing $\mathcal{Q}_2$ on the solution of Problem \ref{QPS3VMProb}, we will start by plotting $z=(p_j+p_{j'}-2p_jp_{j'})$, for all $p_j,p_{j'}\in[0,1]$, as shown in Fig.\ref{z}.

\begin{figure}[h]
%\vspace{-0.15in}
\centering
\includegraphics[width=2.9 in]{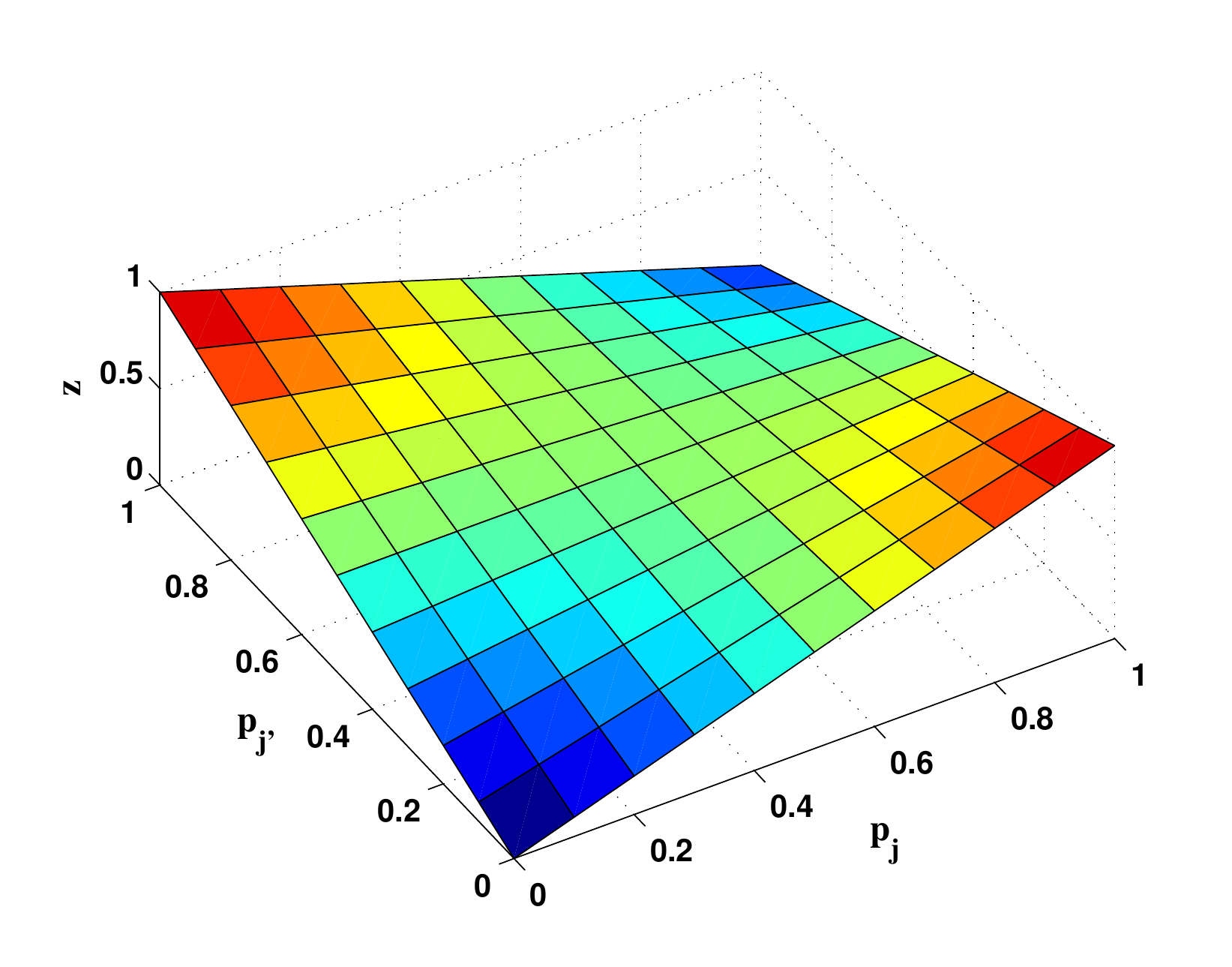}
\vspace{-0.15in}
\caption{Plot of $z=(p_j+p_{j'}-2p_jp_{j'}) \text{ for all } p_j,p_{j'}\in[0,1].$}
\label{z}
\vspace{-0.15in}
\end{figure}

%\vspace{-0.3in}
In Fig.\ref{z} we see that small values of $z$, i.e. $z\simeq 0$, means that $q_j\simeq q_{j'}$ while large values of $z$, i.e. $z\simeq 1$, means that $q_jq_{j'}\simeq 0$. To minimize $\mathcal{Q}_2$ we assign small $z$ to large valued $[{\bf{K}_{uu}}]_{j,j'}$. This means that when two {\em{unlabeled samples}} ${{\bf{x}}_j}$ and ${{\bf{x}}_{j'}}$ are close, $[{\bf{K}_{uu}}]_{j,j'}$ is large, the assigned small valued $z$ will force them to assume the same label, i.e. $q_j\simeq q_{j'}$. On the other hand, if $[{\bf{K}_{uu}}]_{j,j'}$ is small, we assign a large $z$ to it. In other words, if the two unlabeled samples are not close, small $[{\bf{K}_{uu}}]_{j,j'}$, then they should be assigned to different classes, by setting $z$ to be large, i.e. $q_jq_{j'}\simeq 0$. It is easy to see now how minimizing $\mathcal{Q}_2$ basically implements the {\em{clustering assumption}} of semi-supervised learning algorithms where unlabeled samples form clusters and all samples in the same cluster have the same label. Notice that during the minimization of $\mathcal{Q}_2$ a smaller minimum value is achievable if all the unlabeled samples are assigned the same label, that is when $z=0$ and therefore $p_j=p_{j'}$. However, this is a degenerate solution and this is why the balancing constraint is important in the approximate formulation in Problem \ref{QPS3VMProb}

Next we study the second term in Eqn.\eqref{QP-S$^3$VMObjective}. We start by rewriting it as follows: 
\begin{equation}
\label{QPS3VM_Interpret_First_Term}
\begin{aligned}
&CC^*{\bf{Y}}'{\bf{K}_{lu}}({\bf{1}_{\mathcal{|U|}}}-{\bf{P}})
=CC^*\!\!\!\!\!\!\!\stackrel[i\in\mathcal{L},j\in\mathcal{U}]{}{\sum}\!\!\!\!\!\!y_i[{\bf{K}_{lu}}]_{i,j}(1-p_j)
\\
=&{\stackrel[\mathcal{Q}_3]{}{\underbrace{CC^*\!\!\!\!\!\stackrel[\substack{i\in\mathcal{L},j\in\mathcal{U}\\y_i=+1}]{}{\sum}\!\!\!\!\!\![{\bf{K}_{lu}}]_{i,j}(1-p_j)}}}~+\stackrel[\mathcal{Q}_4]{}{\underbrace{CC^*\!\!\!\!\!\stackrel[\substack{i\in\mathcal{L},j\in\mathcal{U}\\y_i=-1}]{}{\sum}\!\!\!\!\!\![{\bf{K}_{lu}}]_{i,j}(p_j-1)}}
\end{aligned}
\end{equation}
We split Eqn \eqref{QPS3VM_Interpret_First_Term} into terms associated with labeled samples with $y_i=+1$, $\mathcal{Q}_3$, and those with $y_i=-1$, $\mathcal{Q}_4$. This is necessary because of the dependence of the interpretation on the labels $y_i$. Since $p_j\in[0,1]$, minimizing $\mathcal{Q}_3$ involves assigning small $(1-p_j)$, i.e. $p_j\simeq 1$, to $[{\bf{K}_{lu}}]_{i,j}$ with large values and vice versa, small valued $[{\bf{K}_{lu}}]_{i,j}$ are assigned large $(1-p_j)$, i.e. $p_j\simeq 0$. In other words, if an {\em{unlabeled sample}} ${\bf{x}}_j$ that is close to, i.e. large $[{\bf{K}_{lu}}]_{i,j}$, a {\em{labeled sample}} $({\bf{x}}_i,y_i=+1)$, then this unlabeled sample should have the same label as the labeled sample, that is $p_j\simeq 1$ and $y_j=+1$. On the other hand, if the {\em{unlabeled sample}} ${\bf{x}}_j$ is far from, i.e. small $[{\bf{K}_{lu}}]_{i,j}$, the {\em{labeled sample}} $({\bf{x}}_i,y_i=+1)$, then this unlabeled sample should have a opposite label to that of the labeled sample, that is $p_j\simeq 0$ and $y_j=-1$. Once again it is notable that if the balancing constraint is not used, a smaller value for the minimum of $\mathcal{Q}_3$ is achievable if all the unlabeled samples are assigned the same label, $p_j=1$ and $y_j=+1$. The same argument holds for minimizing $\mathcal{Q}_4$ where unlabeled samples with large/small similarity to a labeled sample $({\bf{x}}_i,y_i=-1)$ will be assigned small/large $(p_j-1)$, i.e. $p_j\simeq 0$ and $p_j\simeq 1$, respectively.

The process of jointly minimizing $\mathcal{Q}_2$, which implements the clustering assumption of semi-supervised learning, and $\mathcal{Q}_3+\mathcal{Q}_4$, where unlabeled samples are assigned labels by their similarity to labeled samples, results in a formulation that follows the same intuition behind {\em{label propagation}} algorithms \cite{SSLHarmonic} for semi-supervised learning. That is the labeling process chooses dense regions to propagate labels through the unlabeled samples. Therefore, the provided approximate formulation in Problem \ref{QPS3VMProb} does not deviate from the general paradigm of the semi-supervised learning problem. Meanwhile the provided formulation provides an insight into the connection between the {\em{Avoiding Dense Regions}} semi-supervised algorithms, which include S$^3$VM, and the {\em{Graph-based}} algorithms.% {\bf{I THINK THIS PART SHOULD BE ELABORATED UPON. I GUESS WE NEED TO MENTION SOME SPECIFICS, FOR EXAMPLE WITH THE PAPER "ON THE RELATION BETWEEN SPECTRAL CLUSTERING ...ETC."}}.

\section{Submodular Optimization of Approximate QP-S$^3$VM}
\label{Sec:SubmodOpt}
The approximate QP-S$^3$VM formulation proposed in Problem \ref{QPS3VMProb} is simple and intuitive. However, due to the fact that it is a quadratic minimization of a concave function, the computational complexity of finding a solution will become a hindering issue specially for semi-supervised learning problems which are inherently large scale. In this section we use the concepts of submodular set functions to provide a simple and efficient algorithm for the proposed approximate QP-S$^3$VM problem.

Submodular set functions play a central role in combinatorial optimization \cite{GroetschelLovaszSchrijver1993}. They are considered discrete analog of convex functions in continuous optimization in the sense of structural properties that can be benefited from algorithmically. They also emerge as a natural structural form in classic combinatorial problems such as maximum coverage and maximum facility location in location analysis, as well as max-cut problems in graphs. More recently submodular set functions have become key concepts in machine learning where problems such as feature selection \cite {submodularsupermodular} and active learning \cite{NonmyopActive} are solved by maximizing submodular set functions while other core problems like clustering and learning structures of graphical models have been formulated as submodular set function minimization \cite{QClustering}. 

%\subsection{Semi-supervised Learning as a Set Function Optimization Problem}
As discussed in Section \ref{Sec:QP-S$^3$VM} the solution of the approximate QP-S$^3$VM provides a value for the variable $p_j$ associated with each unlabeled sample ${\bf{x}}_j,j\in\mathcal{U}$ such that $p_j=1$ for $y_j=+1$ and $p_j=0$ for $y_j=-1$. In this section we use a different perspective of the problem. In this new perspective the problem of binary semi-supervised classification in general is concerned with choosing a subset $\mathcal{A}$ from the pool of all unlabeled samples $\mathcal{U}$. All the unlabeled samples ${\bf{x}}_j,j\in\mathcal{A}$ should be assigned the label $y_j=+1$ and the rest of them, ${\bf{x}}_j,j\in\mathcal{U}\backslash\mathcal{A}$, will be assigned the label $y_j=-1$. Each possible subset $\mathcal{A}$ is assigned a value by a {\em{set function}} $f(\mathcal{A})$ that has the same optimal solution, in terms of $\mathcal{A}$ and $\mathcal{U}\backslash\mathcal{A}$, as the original semi-supervised classification problem. What makes the reformulation of semi-supervised learning into a set functions interesting is that if the set function $f(\mathcal{A})$ is monotonic {\em{submodular}}, many algorithms can solve the problem efficiently \cite{GroetschelLovaszSchrijver1993}. In the following we give some background on the concept of submodularity in set functions and how we employ it to solve our problem efficiently. 

Let $f({\bf{X}})$ be a set function defined of the set ${\bf{X}}=\{{\bf{x}}_1,{\bf{x}}_2,\dots,{\bf{x}}_n\}$. The monotonicity and submodularity of $f({\bf{X}})$ are defined as follows \cite{GroetschelLovaszSchrijver1993}:
\begin{mydef}%(Diminishing Returns Definition of Submodular Functions)\\
\label{Monoton_Submod_Def}
For all sets $A,B\subseteq\bf{X}$ with $A\subseteq B$, a \textit{set function} $f:2^{{\bf{X}}}\rightarrow\mathbb{R}$ is:\\
a) {{Monotonic}} if
\begin{equation*}
f(A)\leq f(B)
\end{equation*}
b) {{Submodular}} if
\begin{equation*}
f(A\cup\{{\bf{x}}_j\})-f(A)\geq f(B\cup\{{\bf{x}}_j\})-f(B)
\end{equation*}
~~~for all ${\bf{x}}_j\notin B$.
%\vspace{-0.1in}
\end{mydef}
%\etal{GroetschelLovaszSchrijver1993}
A well acknowledged result by Nemhauser {\em{et al.}} \cite{Nemhauser1}, see Theorem \ref{GreedyLowerBoundThm} below, establishes a lower bound of the performance for the simple greedy algorithm, see Algorithm \ref{GreedyAlgorithm}, if it is used to maximize a {\em{monotone submodular set function}} subject to a cardinality constraint. The simple greedy algorithms basically works by adding the element that maximally increases the objective value and according to Theorem \ref{GreedyLowerBoundThm} this simple procedure is guaranteed to achieve at least a constant fraction $(1-1/e)$ of the optimal solution, where $e$ is the natural exponential.% Remember that the provided constant fraction $(1-1/e)$ is just a lower bound on the performance. However, in practice the greedy algorithm achieves significantly better than this lower bound.
\begin{thm2}
\label{GreedyLowerBoundThm}
Given a finite set ${\bf{X}}=\{{\bf{x}}_1,{\bf{x}}_2,\dots,{\bf{x}}_n\}$ and a monotonic submodular function $f(\mathcal{A})$, where $\mathcal{A}\subseteq{\bf{X}}$ and $f(\emptyset)=0$. For the following maximization problem,
\begin{center}
$\mathcal{A}^*=\mathop{argmax}\limits_{\mid \mathcal{A}\mid\leq k}f(\mathcal{A})$.
\end{center}
The greedy maximization algorithm returns $\mathcal{A}_{Greedy}$ such that
\begin{center}
$f(\mathcal{A}_{Greedy})\geq (1-\frac{1}{e})f(\mathcal{A}^*)$.
\end{center}
\end{thm2}}
\floatstyle{ruled}
\newfloat{algorithm}{htbp}{loa}
\floatname{algorithm}{Algorithm}
\begin{algorithm}
\caption{:Greedy Algorithm for Submodular Function Maximization with Cardinality Constraint \cite{Nemhauser1,GreedConstrMaxSub}}
\label{GreedyAlgorithm}
\begin{tabbing}
\=1. Start with ${\bf{X}}_0=\phi$ \\
\>2. For \=$i=1$ to $k$\\
\>\> ${\bf{x}}^*:=argmax_{{\bf{x}}}$  $f({\bf{X}}_{i-1}\cup\{\bf{x}\})-\textit{f }({\bf{X}}_{i-1})$\\
\>\> ${\bf{X}}_i:={\bf{X}}_{i-1}\cup\{{\bf{x}}^*\}$
\end{tabbing}
\end{algorithm}
\subsection{Solving QP-S$^3$VM Using Submodular Optimization}
In this section we use the concepts of submodular functions maximization to provide an efficient and simple algorithm for solving the approximate QP-S$^3$VM problem. Towards this goal we propose the following submodular maximization problem that is equivalent to the approximate QP-S$^3$VM in Problem \ref{QPS3VMProb}.

\begin{myprob}
\label{SubdularApproxProb}
Submodular maximization formulation that is equivalent to Problem \ref{QPS3VMProb}:
\begin{equation}
\stackrel[|\mathcal{A}|\leq r|\mathcal{U}|]{}{max }\mathcal{S}(\mathcal{A})
\end{equation}
\begin{equation}
\label{S(A)Definition}
\begin{aligned}
\text{where     }~~&\\
\mathcal{S}(\mathcal{A})=&-\frac{1}{2}{C^*}^2\!\!\!\!\!\!\! \stackrel[j\in \mathcal{A},j'\in\mathcal{U}]{}{\sum}\!\!\!\!\!\!\left[{\bf{K}_{uu}}\right]_{j,j'}+CC^*\!\!\!\!\!\!\!\stackrel[j\in \mathcal{A},i\in\mathcal{L}]{}{\sum}\!\!\!\!\!\!y_i\left[{\bf{K}_{lu}}\right]_{i,j}
\\&+\frac{1}{2}{C^*}^2\stackrel[j,j'\in\mathcal{A}]{}{\sum}\left[{\bf{K}_{uu}}\right]_{j,j'}
\\&+\!\!\!\stackrel[\mathcal{Q}_5]{}{\underbrace{\stackrel[j,j'\in\mathcal{A}]{}{d \sum}\!\!\left[\delta_{j,j'}\left(\frac{3}{2}{C^*}^2|\mathcal{U}|+CC^*|\mathcal{L}|\right)-\frac{1}{2}{C^*}^2\right]}}
,\end{aligned}
\end{equation}
where $\mathcal{S}$ is a submodular set function defined on all subsets $\mathcal{A}\subset\mathcal{U}$ of unlabeled samples assigned to the class $y_j=+1$, $0\le {\bf{K_{ij}}} \le d$, and  $\delta_{j,j'}=1$ for $j=j'$ and $0$ otherwise.  
\end{myprob}
Problem \ref{SubdularApproxProb} basically maximizes the negative of a discrete version of the objective function in Eqn.\eqref{QP-S$^3$VMObjectiveStandard}. The correspondence between the first three terms in $\mathcal{S(A)}$ and Eqn.\eqref{QP-S$^3$VMObjectiveStandard} is straightforward. However, the term $\mathcal{Q}_5$ is of our design and it is added to ensure the monotonicity and submodularity of $\mathcal{S(A)}$, as shown in Theorem \ref{SubMod_Monoton_Thm}. The constant $d$ is the maximum value of the kernel matrix. Therefore $d=1$ for Radial Basis Function (RBF) kernels. If the data is feature-wise normalized, a highly recommended practice, with values $\in[0,1]$, then for the linear kernel $d$ is equal to the number of dimensions of the used data set (for dense data) or the average number of non-zero features (for sparse data). Since for a fixed $|\mathcal{A}|$ the value of $\mathcal{Q}_5$ is constant, then the optimal solution obtained by optimizing $\mathcal{S(A)}$ is not affected by adding $\mathcal{Q}_5$. In other words $\mathcal{Q}_5$ depends on the cardinality of $\mathcal{A}$ not its contents. 
%Proving that $\stackrel[j,j'\in\mathcal{A}]{}{\sum}\left[\delta_{j,j'}\left(\frac{3}{2}{C^*}^2|\mathcal{U}|+CC^*|\mathcal{L}|\right)-\frac{{C^*}^2}{2}\right]$ is independent from the choice of $\mathcal{A}$.
%\begin{equation*}
%\begin{aligned}
%&\stackrel[j,j'\in\mathcal{A}]{}{\sum}\left[\delta_{j,j'}\left(\frac{3}{2}{C^*}^2|\mathcal{U}|+CC^*|\mathcal{L}|\right)-\frac{{C^*}^2}{2}\right]\\
%&=\left(\frac{3}{2}{C^*}^2|\mathcal{U}|+CC^*|\mathcal{L}|\right)|\mathcal{A}|-\frac{{C^*}^2}{2}{|\mathcal{A}|}^2
%\end{aligned}
%\end{equation*}
%Therefore, for constant $|\mathcal{A}|$, the rest of the expression is constant as well.
\begin{thm2}
\label{SubMod_Monoton_Thm}
The set function $\mathcal{S(A)}$ in Problem \ref{SubdularApproxProb} is monotone (non-decreasing),  submodular, and $\mathcal{S(\emptyset)}=0$. 
\end{thm2} 
\begin{proof} See the appendix.
\end{proof}

Now that we have shown that $\mathcal{S(A)}$ is monotonic, submodular, and $\mathcal{S(\emptyset)}=0$
this means that the greedy maximization algorithm can used be used to optimize Problem \ref{SubdularApproxProb} and the performance guarantee in Theorem \ref{GreedyLowerBoundThm} holds true.

To summarize, the proposed equivalent submodular maximization in Problem \ref{SubdularApproxProb} is defined on the all subsets $\mathcal{A}$ of samples belonging to the class labeled $y_j=+1$. The efficient greedy algorithm in Algorithm \ref{GreedyAlgorithm} is used to the solve the problem efficiently. Once the optimum solution $A^*$ is determined, the rest of the unlabeled samples, i.e. $\mathcal{U}\backslash\mathcal{A}^*$, will belong to class with labels $y_j=-1$. We use the proposed algorithm in the transductive setting of semi-supervised learning. However, if the inductive setting is needed, a standard supervised SVM training can be performed to give the final hyperplane $\bf{w}$.

\section{Experimental Results}
\label{Sec:Exp}
In this section we illustrate the accuracy and efficiency of the proposed QP-S$^3$VM and its submodular optimization (S-QP-S$^3$VM). To this end, we compare the performance of QP-S$^3$VM and S-QP-S$^3$VM with three competitive S$^3$VM algorithms, namely the {\em{Transductive Support Vector Machine}} (TSVM) \cite{citeulike:384511}, the {\em{Deterministic Annealing for Semi-supervised Kernel Machines}} (DA) \cite{DBLP:conf/icml/SindhwaniKC06}, and $\bigtriangledown$TSVM \cite{Chapelle05}. All experiments are performed on a 2 GHZ Intel Core2 Duo machine with 2 GB RAM. The experiments are performed on several real world data, see Table \ref{DatasetTable}, that are selected so as to achieve diversity in terms of dimensionality and distribution properties.    

\begin{table}[h]
%\vspace{-0.1in}
\caption{ Data sets used in the experiments \cite{UCI,cwh03a}.}
\centering
%\hspace{-0.25in}
%\vspace{-0.1in}
\begin{tabular}{l c c c c c c}
\toprule
Data set & Features &  Samples & Labeled & $C$ & $C^*/C$ & $r$ \\
\midrule
australian 		& 14			 & 690 	  	& 3 	 	& 0.922	& 10$^{-1}$	& 0.44\\
w6a 				& 300		 & 1,900	 	& 19		 & 0.838	& 10$^{-4}$	& 0.5\\
svmguide1 		& 4			 & 3,089	 	& 15 	 	& 1.055	& 10$^{-3}$	& 0.65\\
%fourclass 		& 2			 & 862	  	& 4 	 	& 1.276	& 10$^{-2}$	& 0.5\\
a9a 				& 123		 & 15,680	 	& 78		& 0.897	& 10$^{-3}$	& 0.5\\
news20.binary 		& 1,355,191 	& 19,900	 	& 100 	& 6.087	& 10$^{-3}$	& 0.5\\
real-sim 			& 20,958		& 72,309	 	& 8	 	& 1	 	& 10$^{-4}$	& 0.31\\
KDD-99			& 122		& 10$^6$		& 10		& 1		& 10$^{-4}$	& 0.56\\
\bottomrule
\end{tabular}
\label{DatasetTable}
\end{table}

\begin{table*}[htpt]
%\vspace{-0.1in}
\centering
%\hspace{-0.25in}
%\vspace{-0.1in}
\caption{ Classification accuracy experiments for medium size data sets.}
\begin{tabular}{l c c c c c c}
\toprule
Data set & SVM & TSVM & DA & $\bigtriangledown$TSVM & QP-S$^3$VM & S-QP-S$^3$VM  \\
\midrule
australian            & 50.029		& 63.26	 	& 60.48	 	& 56.53 	& {75.57} 	& 74.49\\
w6a                      & 67.44	 	& 58.73	 	& 68.09 		& 52.60 	& {72.33} 	& 70.75\\
svmguide1          & 71.19 		& 77.31 		& 80.98	  	& 69.71 	& {92.73} 	& 92.45\\
%fourclass          &62.228	 	& 65.31 		& 69.99	 	& 59.73 	& {68.37} 	& 69.22\\
a9a 			  & 66.91 		& 71.49		& 72.91		& 64.43 	& -		& 74.90\\
news20.binary   & 63.35	  	& -			& 67.94		& - 		& -		& 71.44 \\
real-sim		  & 52.13		& -			& 69.23		& -		& -		& 71.83\\
KDD-99		  & 72.12		& -			& 97.12		& -		& -		& 98.46\\
\bottomrule
\end{tabular}
\label{AccuracyTable}
\end{table*}

In the accuracy of transductive learning experiment we considered a challenging setup where the number of labeled samples does not exceed 1\% of the available unlabeled data and in two data sets the percentage is as low as 0.01\%. The labeled/unlabeled samples splitting process is repeated 10 times and the average is reported in Table \ref{AccuracyTable}. To illustrate the value of using unlabeled samples in the semi-supervised setting the results of standard SVM trained using only the labeled samples are presented. All experiments use the linear kernel with feature-wise normalized data. The ratio of positive samples in the output $r$ is set to the correct ratio in the unlabeled samples. It is clear in Table \ref{AccuracyTable} that the QP-S$^3$VM and S-QP-S$^3$VM are superior in terms of accuracy to TSVM, DA, and $\bigtriangledown$TSVM. 

In Table \ref{TimeTable} we provide a CPU-time comparison between the QP-S$^3$VM, S-QP-S$^3$VM, TSVM, DA, and $\bigtriangledown$TSVM. It is clear that from the time complexity perspective, S-QP-S$^3$VM is far more efficient than its competitors.

\begin{table}[htpt]
%\vspace{-0.1in}
\centering
%\hspace{-0.25in}
%\vspace{-0.1in}
\caption{ CPU time (Seconds) experiments.}
\begin{tabular}{l c c c c c }
\toprule
Data set  & TSVM & DA & $\bigtriangledown$TSVM & QP-S$^3$VM & S-QP-S$^3$VM  \\
\midrule
australian			& 11.73	 	& 0.786		& 0.452		 & {174.82}	 & 0.013\\
w6a				& 109.40		 & 0.836		& 2.491		 & {6,993.12}	 & 0.038\\
svmguide1		& 186.59	 	& 2.46	 	& 0.803		 & {-}			 & 0.008\\
%fourclass 		& 54.48	 	& 0.408		& 0.117		 & \bf{364.82}	 & 0.013\\
a9a 				& 206.30	 	& 20.78	 	& 18.68		 & -			 & 0.335\\
news20.binary 		& -		 	& 653.4		& -			 & -			 & 3.241 \\
real-sim		 	& -			& 89.38		& -			 & -			 & 1.925\\
KDD-99	     	 	& -			& 2,740		& -			 & -			 & 1,620\\
\bottomrule
\end{tabular}
\label{TimeTable}
\end{table}

\section{Conclusion And Future Work} 
\label{Sec:Conclusion}
In this paper we propose a quadratic programming approximation of the semi-supervised SVM problem (QP-S$^3$VM) that proved to be efficient to solve using standard optimization techniques. One major contribution of the proposed QP-S$^3$VM is that it establishes a link between the two major paradigms of semi-supervised learning, namely low density separation methods and graph-based methods. Such link is considered a significant step towards a unifying framework for semi-supervised learning methods. Furthermore, we propose a novel formulation of the semi-supervised learning problems in terms of submodular set functions which is, up to the authors knowledge, is the first time such idea is presented. Using this new formulation we present a methodology to use submodular optimization techniques to efficiently solve the proposed QP-S$^3$VM problem. Finally, our idea of representing semi-supervised learning problems as submodular set functions will have a great impact on many learning schemes as it will open the door for using an arsenal of algorithms that have theoretical guarantees and efficient performance. The authors are already making progress in extending the presented work to multi-class semi-supervised formulations as well as examining the relationship between submodular optimization over different matroids and its interpretation in terms of semi-supervised learning. One last intriguing point about the proposed work is that samples are assigned to classes, in our case the positive class, sequentially. This opens the door for possible ways to estimate the ratio of positive samples $r$ automatically during the learning process which is still a problem for most semi-supervised techniques specially if there exists a difference in the ratio $r$ between the labeled and unlabeled samples.

\section{Appendix}
\subsection{Proof of Theorem \ref{UpperBoundThm}}
%\begin{proof}
To get an upper bound for ${\bf{\mathcal{I}}_{Dual}}$ we divide it into several components as follows: 
\begin{equation}
{\bf{\mathcal{I}}_{Dual}}={\bf{\mathcal{N}}_1}+{\bf{\mathcal{N}}_2}+{\bf{\mathcal{N}}_3}
%+\\
%
%
\end{equation}
%where
\begin{equation}
\begin{aligned}
\text{where}~~~~&{\bf{\mathcal{N}}_1}={\bf{A'1}}_{|\mathcal{L}|}-\frac{1}{2}({\bf{A}}\circ {\bf{Y}})'{\bf{K}_{ll}}({\bf{A}}\circ {\bf{Y}})\\
&{\bf{\mathcal{N}}_2}=({\bf{\Gamma}}+{\bf{B}})'{\bf{1}}_{|\mathcal{U}|}-\frac{1}{2}({\bf{\Gamma}}-{\bf{B}})'{\bf{K}_{uu}}({\bf{\Gamma}}-{\bf{B}})\\
&{\bf{\mathcal{N}}_3}=-({\bf{A}}\circ{\bf{Y}})'{\bf{K}_{lu}}({\bf{\Gamma}}-{\bf{B}}).
\end{aligned}
\end{equation}
Then
\begin{equation}
%\begin{aligned}
%&\stackrel[\substack{{\bf{0}}\leq{\bf{A}}\leq C{\bf{1}}_{|\mathcal{L}|}\\{\bf{0}}\leq{\bf{\Gamma}}\leq C^*{\bf{P}}\\{{\bf{0}}\leq{\bf{B}}\leq C^*({\bf{1}}_{|\mathcal{U}|}-{\bf{P}})}}]{}{max}{\bf{\mathcal{I}}_{Dual}}\leq\\
%&\stackrel[{\bf{0}}\leq{\bf{A}}\leq C{\bf{1}}_{|\mathcal{L}|}]{}{max}{\bf{\mathcal{N}}_1}+
%\stackrel[\substack{{\bf{0}}\leq{\bf{\Gamma}}\leq C^*{\bf{P}}\\{{\bf{0}}\leq{\bf{B}}\leq C^*({\bf{1}}_{|\mathcal{U}|}-{\bf{P}})}}]{}{max}{\bf{\mathcal{N}}_2}
%+\stackrel[\substack{{\bf{0}}\leq{\bf{A}}\leq C{\bf{1}}_{|\mathcal{L}|}\\{\bf{0}}\leq{\bf{\Gamma}}\leq C^*{\bf{P}}\\{{\bf{0}}\leq{\bf{B}}\leq C^*({\bf{1}}_{|\mathcal{U}|}-{\bf{P}})}}]{}{max}{\bf{\mathcal{N}}_3}
%\end{aligned}
\stackrel[{\bf{A}},{\bf{B}},{\bf{\Gamma}}]{}{max}{\bf{\mathcal{I}}_{Dual}}\leq\\
\stackrel[{\bf{A}}]{}{max}{\bf{\mathcal{N}}_1}+
\stackrel[{\bf{B}},{\bf{\Gamma}}]{}{max}{\bf{\mathcal{N}}_2}
+\stackrel[{\bf{A}},{\bf{B}},{\bf{\Gamma}}]{}{max}{\bf{\mathcal{N}}_3}
\end{equation}
%\begin{equation*}
$\stackrel[{\bf{A}}]{}{max}{\bf{\mathcal{N}}_1}$ 
%\end{equation*}
is the dual form of a standard supervised SVM problem using the label data, i.e. 
\begin{equation}
\stackrel[{\bf{A}}]{}{max}{\bf{\mathcal{N}}_1}=\stackrel[{\bf{w}}]{}{min}\frac{1}{2}\|{\bf{w}}\|^2+C\sum_{i\in\mathcal{L}}\zeta_i
\end{equation}
Furthermore, using the value limits of ${\bf{A}}$, ${\bf{B}}$ and ${\bf{\Gamma}}$, i.e. ${\bf{0}}\leq{\bf{A}}\leq C{{\bf{1}}_{|\mathcal{L}|}}$, ${{\bf{0}}\leq{\bf{B}}\leq C^*({\bf{1}}_{|\mathcal{U}|}-{\bf{P}})}$ and ${\bf{0}}\leq{\bf{\Gamma}}\leq C^*{\bf{P}}$, we can derive the following upper bounds of ${\bf{\mathcal{N}}_2}$ and ${\bf{\mathcal{N}}_3}$, 
\begin{equation}
\stackrel[{\bf{B}},{\bf{\Gamma}}]{}{max}{\bf{\mathcal{N}}_2}\leq C^*|\mathcal{U}|+\frac{1}{2}{C^*}^2({\bf{1}_{\mathcal{|U|}}}-{\bf{P}})'{\bf{K}_{uu}}{\bf{P}}
\end{equation}
and
\begin{equation}
\stackrel[{\bf{A}},{\bf{B}},{\bf{\Gamma}}]{}{max}{\bf{\mathcal{N}}_3}\leq CC^*{\bf{Y}}'{\bf{K}_{lu}}({\bf{1}_{\mathcal{|U|}}}-{\bf{P}}).
\end{equation}
Combining the three upper bounds we get the provided bound in the theorem.
%\end{proof}

\subsection{Proof of Theorem \ref{SubMod_Monoton_Thm}}
%\begin{proof}
%\begin{mydef}
%\label{}
%Proof of the non-decreasing property of $\mathcal{S}(\mathcal{A)}$.\\
%\begin{equation*}
%\begin{aligned}
%&\mathcal{S(A)}=\frac{{-C^*}^2}{2} \stackrel[j\in \mathcal{A},j'\in\mathcal{U}]{}{\sum}\left[\bf{K}_{u,u}\right]_{j,j'}+CC^*\stackrel[j\in \mathcal{A},i\in\mathcal{L}]{}{\sum}y_i\left[\bf{K}_{l,u}\right]_{i,j}+\frac{{C^*}^2}{2}\stackrel[j,j'\in\mathcal{A}]{}{\sum}\left[\bf{K}_{u,u}\right]_{j,j'}\\
%&+\stackrel[j,j'\in\mathcal{A}]{}{\sum}\left[\delta_{j,j'}\left(\frac{3}{2}{C^*}^2|\mathcal{U}|+CC^*|\mathcal{L}|\right)-\frac{{C^*}^2}{2}\right]
%\end{aligned}
%\end{equation*}
First, $\mathcal{S(\emptyset)}=0$ follows directly from the definition in Eqn.\eqref{S(A)Definition} where all the summations are on elements in the set $\mathcal{A}$. Therefore if $\mathcal{A}=\emptyset$ then $\mathcal{S(\emptyset)}=0$. For the sake of simplicity we consider the special case where $d=1$. However, the extension to the general values of $d$ is fairly straightforward. Next we prove the {\em{monotonicity property}}. Using the definition of $\mathcal{S}(\mathcal{A)}$, we can show that for any $m\in\mathcal{U}$ and $m\notin\mathcal{A}$, the increase in the objective value of $\mathcal{S}$ due to adding $m$ is,
%\begin{equation*}
%\begin{aligned}
%&\text{then  }\\
%\mathcal{S}(\mathcal{A}\!\cup\! m)\!=\!&-\frac{1}{2}{C^*}^2\!\!\!\!\!\!\!\!\!\!\!\! \stackrel[j\in\{\mathcal{A}\cup m\},j'\in\mathcal{U}]{}{\sum}\!\!\!\!\!\!\!\!\!\!\!\!\left[{\bf{K}_{uu}}\right]_{j,j'}\!+CC^*\!\!\!\!\!\!\!\!\!\!\!\!\stackrel[j\in\{\mathcal{A}\cup m\},i\in\mathcal{L}]{}{\sum}\!\!\!\!\!\!\!\!\!\!\!\!y_i\left[{\bf{K}_{lu}}\right]_{i,j}
%\\&+\frac{1}{2}{C^*}^2\!\!\!\!\!\!\!\!\stackrel[j,j'\in\{\mathcal{A}\cup m\}]{}{\sum}\!\!\!\!\!\left[{\bf{K}_{uu}}\right]_{j,j'}\\
%&+\!\!\!\!\!\!\!\!\!\!\stackrel[j,j'\in\{\mathcal{A}\cup m\}]{}{\sum}\!\!\!\left[\delta_{j,j'}\!\!\left(\frac{3}{2}{C^*}^2|\mathcal{U}|\!+\!CC^*|\mathcal{L}|\right)\!-\!\frac{1}{2}{C^*}^2\right]
%\end{aligned}
%\end{equation*}
%\begin{equation*}
%\begin{aligned}
%&\text{therefore  }\\
%&\mathcal{S}(\mathcal{A}\cup m)-\mathcal{S}(\mathcal{A})=
%\\&-\frac{1}{2}{C^*}^2\stackrel[j'\in\mathcal{U}]{}{\sum}\left[{\bf{K}_{uu}}\right]_{m,j'}+CC^*\stackrel[i\in\mathcal{L}]{}{\sum}y_i\left[{\bf{K}_{lu}}\right]_{i,m}
%\\&+\frac{1}{2}{C^*}^2\left[2\stackrel[j'\in\mathcal{A}]{}{\sum}\left(\left[{\bf{K}_{uu}}\right]_{m,j'}-1\right)+\left(\left[{\bf{K}_{uu}}\right]_{m,m}-1\right)\right]
%\\&+\frac{3}{2}{C^*}^2|\mathcal{U}|+CC^*|\mathcal{L}|
%\end{aligned}
%\end{equation*}
\begin{equation}
\begin{aligned}
%&\text{that is   }\\
&\mathcal{S}(\mathcal{A}\cup m)-\mathcal{S}(\mathcal{A})=
\\&-\frac{1}{2}{C^*}^2\stackrel[j'\in\mathcal{U}]{}{\sum}\left[{\bf{K}_{uu}}\right]_{m,j'}+CC^*\stackrel[i\in\mathcal{L}]{}{\sum}y_i\left[{\bf{K}_{lu}}\right]_{i,m}
\\&+{C^*}^2\stackrel[j'\in\mathcal{A}]{}{\sum}\left[{\bf{K}_{uu}}\right]_{m,j'}-{C^*}^2|\mathcal{A}|
\\&+\frac{1}{2}{C^*}^2\left(\left[{\bf{K}_{uu}}\right]_{m,m}-1\right)+
\frac{3}{2}{C^*}^2|\mathcal{U}|+CC^*|\mathcal{L}|
\end{aligned}
\end{equation}
Since we are examining the case where $d=1$, then ${0\leq\bf{K}}_{i,j}\leq 1$ and ${\bf{K}}_{i,i}=1$. Therefore, since
\begin{equation}
\begin{aligned}
%\text{since      }\\
&\frac{1}{2}{C^*}^2\left(\left[{\bf{K}_{uu}}\right]_{m,m}-1\right)=0\\
&{C^*}^2\stackrel[j'\in\mathcal{A}]{}{\sum}\left[{\bf{K}_{uu}}\right]_{m,j'}\geq 0\\
%\end{aligned}\\
%&\text{and   }\\
%\begin{aligned}
&CC^*|\mathcal{L}|+ CC^*\stackrel[i\in\mathcal{L}]{}{\sum}y_i\left[{\bf{K}_{lu}}\right]_{i,m}\geq 0\\
%&\text{and   }\\
&\frac{3}{2}{C^*}^2|\mathcal{U}|\geq\frac{1}{2}{C^*}^2\stackrel[j'\in\mathcal{U}]{}{\sum}\left[{\bf{K}_{uu}}\right]_{m,j'}+{C^*}^2|\mathcal{A}|\\
%\text{then   }\\
\end{aligned}
\end{equation}
then
\begin{equation*}
\mathcal{S}(\mathcal{A}\cup m)-\mathcal{S}(\mathcal{A})\geq 0 
\end{equation*}
%\end{mydef}
Thus the monotonicity property of $\mathcal{S(A)}$ holds true.

Now we prove the {\em{submodularity}} of $\mathcal{S(A)}$ by assuming the set $\mathcal{B}=\{\mathcal{A}\cup q\}$ where $q\in\mathcal{U}$. Using the same set element $m$ we used earlier, i.e. $m\in\mathcal{U}$ and $m\notin\mathcal{A}$, we need to show that adding $m$ to the set $\mathcal{A}$ has more effect than adding it to the set $\mathcal{B}$ as stated in Definition \ref{Monoton_Submod_Def}-b. Since 
\begin{equation}
\begin{aligned}
%&\text{Let  } \mathcal{B}=\{\mathcal{A}\cup q\}\\
%&\text{then  }\\
\mathcal{S}(\mathcal{B})=&-\frac{1}{2}{C^*}^2\!\!\!\!\!\!\!\!\!\!\!\!\stackrel[j\in \{\mathcal{A}\cup q\},j'\in\mathcal{U}]{}{\sum}\!\!\!\!\!\!\!\!\!\!\!\!\left[{\bf{K}_{uu}}\right]_{j,j'}\!+\!CC^*\!\!\!\!\!\!\!\!\!\!\!\!\stackrel[j\in \{\mathcal{A}\cup q\},i\in\mathcal{L}]{}{\sum}\!\!\!\!\!\!\!\!\!\!\!y_i\left[{\bf{K}_{lu}}\right]_{i,j}
\\&+\frac{1}{2}{C^*}^2\!\!\!\!\!\!\!\!\!\stackrel[j,j'\in\{\mathcal{A}\cup q\}]{}{\sum}\left[{\bf{K}_{uu}}\right]_{j,j'}\\
&+\!\!\!\!\!\!\!\!\stackrel[j,j'\in\{\mathcal{A}\cup q\}]{}{\sum}\!\!\!\left[\delta_{j,j'}\!\!\left(\frac{3}{2}{C^*}^2|\mathcal{U}|+CC^*|\mathcal{L}|\right)\!-\!\frac{1}{2}{C^*}^2\right]\\
\end{aligned}
\end{equation}
%\begin{equation*}
%\begin{aligned}
%&\text{and}\\
%\mathcal{S}(\mathcal{B}\!\cup\! m)\!=\!&-\frac{1}{2}{C^*}^2\!\!\!\!\!\!\!\!\!\!\! \stackrel[j\in \{\mathcal{A}\cup q\cup m\},j'\in\mathcal{U}]{}{\sum}\!\!\!\!\!\!\!\!\!\!\!\!\!\!\!\left[{\bf{K}_{uu}}\right]_{j,j'}\!+\!CC^*\!\!\!\!\!\!\!\!\!\stackrel[j\in \{\mathcal{A}\cup q\cup m\},i\in\mathcal{L}]{}{\sum}\!\!\!\!\!\!\!\!\!\!\!\!\!\!\!y_i\left[{\bf{K}_{lu}}\right]_{i,j}
%\\&+\frac{1}{2}{C^*}^2\stackrel[j,j'\in\{\mathcal{A}\cup q\cup m\}]{}{\sum}\!\!\!\!\!\!\!\!\left[{\bf{K}_{uu}}\right]_{j,j'}\\
%&+\!\!\!\!\!\!\!\!\!\stackrel[j,j'\in\{\mathcal{A}\cup q\cup m\}]{}{\sum}\!\!\!\left[\delta_{j,j'}\!\!\left(\frac{3}{2}{C^*}^2|\mathcal{U}|\!+\!CC^*|\mathcal{L}|\right)\!-\!\frac{1}{2}{C^*}^2\right]\\
%\end{aligned}
%\end{equation*}
then
\begin{equation}
\begin{aligned}
%&\text{then  }\\
\mathcal{S}(\mathcal{B}&\cup m)-\mathcal{S}(\mathcal{B})=
\\&-\frac{1}{2}{C^*}^2\!\! \stackrel[j'\in\mathcal{U}]{}{\sum}\!\!\left[{\bf{K}_{uu}}\right]_{m,j'}+CC^*\!\stackrel[i\in\mathcal{L}]{}{\sum}\!y_i\left[{\bf{K}_{lu}}\right]_{i,m}\\&+{C^*}^2\stackrel[j'\in\{\mathcal{A}\cup q\}]{}{\sum}\!\!\!\left[{\bf{K}_{uu}}\right]_{m,j'}-{C^*}^2\left(|\mathcal{A}|+1\right)
\\&+\frac{1}{2}{C^*}^2\left(\left[{\bf{K}_{uu}}\right]_{m,m}-1\right)+
\frac{3}{2}{C^*}^2|\mathcal{U}|+CC^*|\mathcal{L}|
\end{aligned}
\end{equation}
Therefore
\begin{equation}
\begin{aligned}
%&\text{Therefore  }\\
&\left(\mathcal{S}(\mathcal{A}\cup m)-\mathcal{S}(\mathcal{A})\right)-\left(\mathcal{S}(\mathcal{B}\cup m)-\mathcal{S}(\mathcal{B})\right)=
\\&~~~~~~~~~~~~~~~~~~~{{C^*}^2}\left(1-\left[\bf{K}_{u,u}\right]_{q,m}\right)\geq 0
\end{aligned}
\end{equation}
Hence the set function $\mathcal{S(A)}$ is submodular.
%\end{proof}

\bibliographystyle{IEEEtran.bst}
\bibliography{ProposalMachLearnBib}

\end{document}